\documentclass[sigconf]{acmart}
\usepackage{pifont}
\usepackage{multirow}
\usepackage{enumitem}
\usepackage{subfigure}
\newtheorem{definition}{Definition}
\newtheorem{theorem}{Theorem}

\newcommand{\nosection}[1]{\vspace{2pt}\noindent\textbf{#1:}}

\AtBeginDocument{%
  \providecommand\BibTeX{{%
    \normalfont B\kern-0.5em{\scshape i\kern-0.25em b}\kern-0.8em\TeX}}}

\copyrightyear{2023}
\acmYear{2023}
\setcopyright{acmlicensed}
\acmConference[MM '23] {Proceedings of the 31st ACM International Conference on Multimedia}{October 29--November 3, 2023}{Ottawa, ON, Canada.}
\acmBooktitle{Proceedings of the 31st ACM International Conference on Multimedia (MM '23), October 29--November 3, 2023, Ottawa, ON, Canada}
\acmPrice{15.00}
\acmISBN{979-8-4007-0108-5/23/10}
\acmDOI{10.1145/3581783.3612418}

\begin{document}

\title{Making Users Indistinguishable: Attribute-wise Unlearning in Recommender Systems}

\author{Yuyuan Li}
\orcid{0000-0003-4896-2885}
\affiliation{%
  \institution{College of Computer Science, Zhejiang University}
  \city{Hangzhou}
  \country{China}
}
\email{11821022@zju.edu.cn}

\author{Chaochao Chen}
\orcid{0000-0003-1419-964X}
\authornote{Corresponding author.}
\affiliation{%
  \institution{College of Computer Science, Zhejiang University}
  \city{Hangzhou}
  \country{China}
}
\email{zjuccc@zju.edu.cn}

\author{Xiaolin Zheng}
\orcid{0000-0001-5483-0366}
\affiliation{%
  \institution{College of Computer Science, Zhejiang University}
  \city{Hangzhou}
  \country{China}
}
\email{xlzheng@zju.edu.cn}

\author{Yizhao Zhang}
\orcid{0009-0008-0241-8706}
\affiliation{%
  \institution{College of Computer Science, Zhejiang University}
  \city{Hangzhou}
  \country{China}
}
\email{22221337@zju.edu.cn}

\author{Zhongxuan Han}
\orcid{0000-0001-9957-7325}
\affiliation{%
  \institution{College of Computer Science, Zhejiang University}
  \city{Hangzhou}
  \country{China}
}
\email{zxhan@zju.edu.cn}

\author{Dan Meng}
\orcid{0000-0003-1980-9283}
\affiliation{%
  \institution{OPPO Research Institute}
  \city{Shenzhen}
  \country{China}
}
\email{mengdan90@163.com}

\author{Jun Wang}
\orcid{0000-0002-0481-5341}
\affiliation{%
  \institution{OPPO Research Institute}
  \city{Shenzhen}
  \country{China}
}
\email{junwang.lu@gmail.com}

\renewcommand{\shortauthors}{Yuyuan Li et al.}

\begin{abstract}
  With the growing privacy concerns in recommender systems, recommendation unlearning, i.e., forgetting the impact of specific learned targets, is getting increasing attention.
  Existing studies predominantly use training data, i.e., model inputs, as the unlearning target. 
  However, we find that attackers can extract private information, i.e., gender, race, and age, from a trained model even if it has not been explicitly encountered during training.
  We name this unseen information as \textit{attribute} and treat it as the unlearning target.
  To protect the sensitive attribute of users, Attribute Unlearning (AU) aims to degrade attacking performance and make target attributes indistinguishable.
  In this paper, we focus on a strict but practical setting of AU, namely Post-Training Attribute Unlearning (PoT-AU), where unlearning can only be performed after the training of the recommendation model is completed. 
  To address the PoT-AU problem in recommender systems, we design a two-component loss function that consists of i) distinguishability loss: making attribute labels indistinguishable from attackers, and ii) regularization loss: preventing drastic changes in the model that result in a negative impact on recommendation performance. 
  Specifically, we investigate two types of distinguishability measurements, i.e., user-to-user and distribution-to-distribution. 
  We use the stochastic gradient descent algorithm to optimize our proposed loss.
  Extensive experiments on three real-world datasets demonstrate the effectiveness of our proposed methods.
\end{abstract}




\maketitle

\section{Introduction}\label{sec:intro}
Recommendation unlearning has gained increasing interest in recent years.
On the one hand, recommender systems have been widely applied in practice with great success, having a substantial influence on people's lifestyles~\cite{schafer2007collaborative, han2023intra, chen2022differential}.
The success lies in their ability to extract highly personalized information from user data.
On the other hand, people have grown more aware of privacy concerns in personalized recommendations, and demand their sensitive information be protected.
As one of the protective measures, \textit{Right to be Forgotten}~\cite{2014gdpr, 2018ccpa, 2019pipeda} requires recommendation platforms to enable users to withdraw their individual data and its impact, which impulses the study of machine/recommendation unlearning.


\begin{table}[t]
    \centering
    \caption{Difference between input unlearning and attribute unlearning in recommender systems.}
    \resizebox{\linewidth}{!}{
    \begin{tabular}{c|c|c}
         \toprule
           &  \textbf{Input Unlearning} & \textbf{Attribute Unlearning}\\
         \midrule
         \multirow{2}{*}{Unlearning target} & Input data & Latent attribute\\
         & (used in training) & (\textbf{not} used in training)\\
         \midrule
         Applicability of & \multirow{2}{*}{Ground truth} & \multirow{2}{*}{Not applicable}\\
         retraining from scratch & \\
         \bottomrule
    \end{tabular}
    }
    \label{tab:type}
\end{table}

Existing studies of machine unlearning mainly use training data, i.e., model inputs, as the unlearning target~\cite{nguyen2022survey}.
We name this type of unlearning task as Input Unlearning (IU). 
%
In the recommendation scenarios, the input data can be a user-item interaction matrix.
With different unlearning targets, IU can be user-wise, item-wise, and instance-wise~\cite{chen2022recommendation}.
IU benefits multiple parties, e.g., data providers and model owners, because the target data can be i) the specified data that contains users' sensitive information, and ii) the dirty data that is polluted by accidental mistakes or intentional attack~\cite{li2016data}.

Extensive studies on IU cannot obscure the importance of Attribute Unlearning (AU), where attributes represent the inherent properties, e.g., gender, race, and age of users that have \textbf{not} been used for training (Table~\ref{tab:type}: difference in unlearning target).
Due to the information extraction capabilities of recommender systems, AU is especially valuable in the context of recommendation.
Although recommendation models did not see the latent attribute, the research found that basic machine learning models can successfully infer users' attributes from the user embedding learned by collaborative filtering models~\cite{ganhor2022unlearning}, which is also known as Attribute Inference Attack (AIA)~\cite{jia2018attriguard, beigi2020privacy}.
Therefore, from the perspective of privacy preservation, AU is as important as IU in recommender systems.
However, existing IU methods cannot be applied in AU.
As illustrated in Table~\ref{tab:type}, retraining from scratch (ground truth for IU) is unable to unlearn the latent attribute, i.e., not applicable for AU, since it is not utilized during training at all.

Existing but limited research on AU has focused on In-Training AU (InT-AU)~\cite{guo2022efficient,ganhor2022unlearning}, where unlearning is performed during model training (as shown in the left of Figure~\ref{fig:overview}).
In this paper, we focus on a more strict AU setting, namely Post-Training Attribute Unlearning (PoT-AU), where we can only manipulate the model, i.e., updating parameters, after the training is fully completed and have no knowledge about training data or other training information (as shown in the right of Figure~\ref{fig:overview}).
Compared with InT-AU, this setting is more practical, because of i) data accessibility: we may not get access to the training data or other information after training due to regulations, and ii) deployment overhead: non-interference with the original training process is more flexible and reduces deployment overhead.
%
%
As shown in Figure~\ref{fig:overview}, there are two goals for PoT-AU in recommender systems, where user embedding is the target of attacking and unlearning.
The primary goal (\textbf{Goal \#1}) is to make the target attribute indistinguishable to AIA, or more directly, to degrade the attacking performance.
The other goal (\textbf{Goal \#2}) is to maintain the recommendation performance.
This goal is equally important as the primary one, since both users and recommendation platforms want to avoid having a negative impact on the original recommendation tasks.

To achieve the above two goals in the PoT-AU problem, we consider it as an optimization problem on user embedding which is the crucial parameter in the recommendation model that contains user information. 
We then design a two-component loss function that consists of i) \textit{distinguishability loss}: measuring the distinguishable degree of users with different attribute labels, and ii) \textit{regularization loss}: measuring the divergence of current parameters and the original ones.
Each component is individually devised for one goal in the PoT-AU problem.
We introduce the general design principle of each component, which is applicable to all recommendation models that have user embeddings. 
Specifically, we mainly focus on the design of distinguishability loss and investigate two types of distinguishability measurements from different perspectives, i.e., User-to-User (U2U) and Distribution-to-Distribution (D2D).
U2U loss measures the distinguishability by the weighted embedding divergence of each user-pairs if the two users in the pair have different attribute labels.
D2D loss regards all users within the same attribute label as a distribution and measures the distinguishability by the distance of different distributions.
We implement our proposed loss functions with effective and computationally efficient components, and adopt stochastic gradient descent to optimize them.

We summarize the main contributions of this paper as follows:
\begin{itemize}[leftmargin=*]\setlength{\itemsep}{-\itemsep}
    \item To the best of our knowledge, we are the first to study the PoT-AU problem in recommender systems, which is a more strict and practical problem than InT-AU.
    We identify two essential goals of PoT-AU in recommender systems, i.e., making attributes indistinguishable, and maintaining recommendation performance.
    \item To address the PoT-AU problem, we propose a two-component loss function, i.e., a linear combination of distinguishability loss and regularization loss.
    Each component is devised to achieve one of the above goals separately.
    From two different perspectives, we design two types of distinguishability loss, i.e., U2U and D2D.
    \item We implement two examples, i.e., U2U-R and D2D-R, of our proposed loss function and conduct extensive experiments on three real-world datasets to comprehensively evaluate the effectiveness of our proposed methods in terms of both unlearning (\textbf{Goal \#1}) and recommendation (\textbf{Goal \#2}).
    \item To better understand the mechanism of our proposed method, we analyze the user embedding changes before and after unlearning, finding the connection between these changes and the performance difference between U2U and D2D.
    Finally, we also discuss the potential variations of different distinguishability losses.
\end{itemize}

\section{Preliminaries}

\subsection{Recommendation Model}\label{sec:rec}
Today's recommender systems can provide personalized recommendations based on collaborative information across users and items.
Collaborative filtering is an acknowledged algorithm for analyzing this information~\cite{shi2014collaborative}.
In this paper, we choose the widely applied collaborative filtering which is based on matrix factorization (MF-based CF).
The basic idea of MF-based CF is to decompose the user-item interaction matrix into two low-rank embedding matrices, i.e., user embedding and item embedding.
In the PoT-AU problem, we use user embedding as the target of attacking and unlearning.

\begin{figure*}[t]
    \centering
    \includegraphics[width=\linewidth]{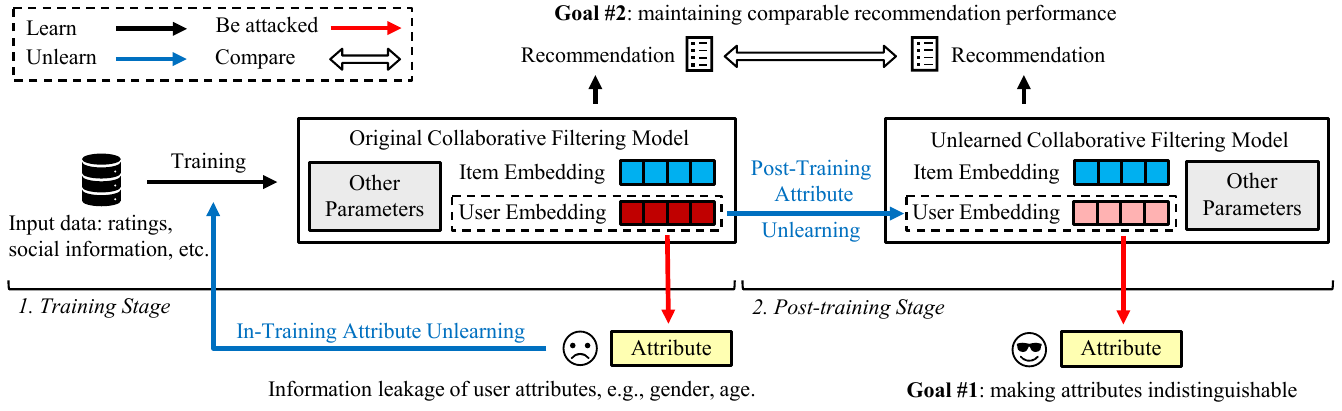}
    \caption{Overview of Post-Training Attribute Unlearning (PoT-AU) in recommender systems.}
    \Description{A flow chart of unlearning and unlearning workflows in recommender systems.}
    \label{fig:overview}
\end{figure*}

\subsection{Attacking Setting}

The process of attacking in PoT-AU problem is also known as AIA, which consists of three stages, i.e., exposure, training, and inference.
%
%
In the exposure stage, we assume that attackers follow the grey-box attack.
In other words, not all model parameters but only users' embedding and their corresponding attribute information are exposed to attackers.
In the training stage, we assume that attackers train the attacking model on a shadow dataset, which can be generated by sampling from the original users or users from the same distribution~\cite{salem2018ml}.
Although shadow-dataset training will inevitably reduce attacking performance, this assumption is reasonable, since the full-dataset setting is too strong and impractical.
Regarding the attack as a classification task, attacking models use user embedding as input data and attribute information as labels.
For conciseness, we focus on binary classification in this paper.
We will discuss the generalization to multi-class classification in Section~\ref{sec:con}.
In the inference stage, attackers use their trained attacking models to make predictions.

\section{Post-Training Attribute Unlearning}
In this section, we first further explain our motivation as well as the process of the PoT-AU problem in recommender systems. 
Then we consider the PoT-AU problem as an optimization problem on user embedding and propose a novel two-component loss function to address it.

\subsection{Motivation}
As shown in Figure~\ref{fig:overview}, we divide the whole process of PoT-AU into two stages, i.e., training stage and post-training stage.
In the training stage, the recommender system trains an original collaborative filtering model on input data.
To align with the post-training setting, we leave this stage untamed and assume that one has no information in this stage except the user embedding and their attributes.
In the post-training stage, we generate new user embedding by unlearning the original one. 
The new embedding, i.e., user embedding after unlearning, is supposed to achieve two goals simultaneously.
\textbf{Goal \#1} (unlearning) is to make target attributes distinguishable so as to protect attribute information from attackers.
\textbf{Goal \#2} (recommendation) is to maintain the original recommendation performance so as not to influence users' initial requirements.

Compared with in-training setting (InT-AU), the post-training setting (PoT-AU) is more challenging.
Firstly, PoT-AU allows no interference with the training process, which means InT-AU methods, i.e., adding network block~\cite{guo2022efficient}, and adversarial training~\cite{ganhor2022unlearning}, are not applicable.
Secondly, even though PoT-AU cuts down the connection with the training process, directly manipulating user embedding by adding artificially-designed noise, e.g., differential privacy~\cite{abadi2016deep}, is inappropriate, because i) it will inevitably degrade recommendation performance, and ii) its unlearning ability is not promising, as the functional mechanism of attacking models, including complex machine learning models, is not well-understood. 
Thirdly, PoT-AU prohibits access to the input data and other training information, as the said data could be either unavailable or under protection.
As a result, the said data cannot be used for fine-tuning user embedding, e.g., adding noise to the embedding and then fine-tuning to boost recommendation performance.
%
%
%
%

\subsection{Two-Component Loss Function}
One of the feasible solutions is to envision the final desired user embedding and temporally ignore the intermediate manipulation and transformation. 
Thus, we regard PoT-AU as an optimization problem on user embedding.
In other words, we aim to design a proper loss function and let optimization techniques do the rest.

There are two goals in PoT-AU problem, i.e., \textbf{Goal \#1}: unlearning and \textbf{Goal \#2}: recommendation.
Thus, we propose a two-component loss function in which one component is individually devised for one goal.
Formally, we combine the two components linearly with a trade-off parameter $\alpha$ to balance both goals: 
\begin{equation}\label{equ:loss}
    L(\theta) = \ell_u + \alpha \ell_r,
\end{equation}
where $\theta \in \mathbb{R}^{N\times K}$ denotes user embedding ($N$ is the number of users, and $K$ is the number of embedding dimensions), $\ell_u$ and $\ell_r$ represent the loss for \textbf{Goal \#1} and \textbf{Goal \#2} respectively.


\subsection{Distinguishability Loss}
The core difficulty of designing a proper two-component loss function lies in defining distinguishability loss $\ell_u$.
It is related to the primary goal of PoT-AU, i.e., \textbf{Goal \#1}. 
%
We define the distinguishability from two perspectives, i.e., User-to-User (U2U) and Distribution-to-Distribution (D2D). 
Following the definitions, we design two types of distinguishability losses respectively.
For conciseness, we assume the target attribute has binary labels: $S_1$ and $S_2$.

\subsubsection{\textbf{User-to-user Loss}} 
The attacker infers users' attributes according to their embedding. 
From a U2U perspective, making attributes indistinguishable means letting users with different attribute labels have similar embedding.
This idea is statistically expressed as a positive correlation between the divergence of attribute information and the similarity of user embedding.
In other words, there is a negative correlation between the divergence of attribute information and the divergence of user embedding, which can be formally expressed as:
\begin{equation}\label{equ:neg}
    Div(\theta_i, \theta_j) \le \frac{\delta}{Div(v_i, v_j)} \text{ for } i\in S_1, j\in S_2,
\end{equation}
where $Div$ can be certain divergence functions, $v_i$ denotes the attribute information of user $i$, and $\delta$ is a constant.
The attribute information is usually regarded as the attribute labels, but there are more advanced options in the PoT-AU setting, which we will discuss in Section~\ref{sec:imp}.

As user embedding $\theta$ is the target to optimize, Equation (\ref{equ:neg}) can be interpreted as the divergence of embedding $Div(\theta_i, \theta_j)$ is bounded by a scale of $\delta/Div(v_i, v_j)$. 
That is, the greater the divergence between the attribute information of users $i$ and $j$, the tighter the upper bound, and hence the smaller the divergence in embedding between users is likely to be.
Assuming that the divergence function is non-negative, we can rewrite Equation (\ref{equ:neg}) as: 
\begin{equation}\label{equ:dis}
    Div(\theta_i, \theta_j) \cdot Div(v_i, v_j) \le \delta.
\end{equation}
Equation (\ref{equ:dis}) reflects the concept of U2U-perspective distinguishability, where $\delta$ is the upper bound of the distinguishability degree.
We name this type of distinguishability measurement as U2U loss and define it as follows.
\begin{definition}[User-to-user distinguishability]
    Given the divergence of user embedding $Div(\theta_i, \theta_j)$ and the divergence of user attribute $Div(v_i, v_j)$, we define user-to-user distinguishability as:
    \begin{equation}\label{equ:u2u_def}
        \ell_{u, U} = \sum_{i \in S_1}\sum_{j \in S_2} Div(\theta_i, \theta_j) \cdot Div(v_i, v_j).
    \end{equation}
\end{definition}
Note that the divergence of embedding can be different from that of attribute, which leads to more possibility of U2U loss.
Interpreting $Div(v_i, v_j)$ as a weight term, Equation (\ref{equ:u2u_def}) shows that U2U loss is a weighted divergence loss of user embedding.
The larger the weight, the more stringent the measure of embedding divergence is likely to be, which reflects our design principle, i.e., letting users with different attribute labels have similar embedding.

\subsubsection{\textbf{Distribution-to-distribution Loss}} We consider the user embedding with the same attribute label as a distribution, e.g., $P_{\theta_1}$ denotes the embedding distribution of users with label $S_1$.
Inspired by Theorem~\ref{the:domain}, the distinguishability of user performance can be bounded by the $\mathcal{H}$-divergence of their distributions $div_{\mathcal{H}}(P_{\theta_1}, P_{\theta_2})$.
\begin{theorem}\label{the:domain}[Bound on Domain Risk~\cite{shar2010atheory}] Let $\mathcal{H}$ be a hypothesis class of VC dimension $d$. With probability $1 - \epsilon$ over the choice of samples $\Theta_1 \sim P(\theta_1)^n$ and $\Theta_2 \sim P(\theta_2)^n$, for every $\eta \in \mathcal{H}$:
\begin{equation}
    R_{\Theta_1}(\eta) - R_{\Theta_2}(\eta) \le \hat{div}_\mathcal{H}(\Theta_1, \Theta_2) + \beta + \gamma,
\end{equation}
with $\beta = \sqrt{\frac{4}{n}(d\log\frac{2en}{d} + \log\frac{4}{\epsilon})} + 4\sqrt{\frac{1}{n}(d\log\frac{2n}{d} + \log\frac{4}{\epsilon})}$, and $\gamma \ge \inf_{\eta^*\in \mathcal{H}}[R_{P_{\theta_1}}(\eta^*) + R_{P_{\theta_2}}(\eta^*)]$.
\end{theorem}
In Theorem~\ref{the:domain}, $\hat{div}_\mathcal{H}$ is the empirical $\mathcal{H}$-divergence, and $R$ is the domain risk which represents the recommendation loss in our scenarios, reflecting the performance of users.
For practical consideration, it is worth noting that the embeddings of all users are trained together without any attribution information. 
As a result, the shapes of the embedding distribution tend to be similar across different attribute labels.
The difference in distributions mainly comes from their distance.
Therefore, we use $dist(P_{\theta_1}, P_{\theta_2})$ to approximate $div_{\mathcal{H}}(P_{\theta_1}, P_{\theta_2})$ and omit the constants for simplicity.
We name this type of distinguishability measurement as D2D loss and define it as follows.
\begin{definition}[Distribution-to-distribution distinguishability]\label{def:d2d}
    Given two distributions of embedding from users with different attribute labels $P_{\theta_1}$ and $P_{\theta_2}$, we define distribution-to-distribution distinguishability as the distance between two distributions:
    \begin{equation}
        \ell_{u, D} = Dist(P_{\theta_1}, P_{\theta_2}).
    \end{equation}
\end{definition}
Thus, distributional distances, e.g., KL divergence~\cite{goldberger2003efficient} and Maximum Mean Discrepancy (MMD)~\cite{smola2006maximum}, can be used to measure the degree of D2D distinguishability.

\subsection{Regularization Loss}
To achieve \textbf{Goal \#2}, we add regularization loss $\ell_r$ in Equation~(\ref{equ:loss}).
We name $\ell_r$ as regularization loss instead of recommendation loss.
The reason is that we cannot use common recommendation loss, e.g., binary cross-entropy~\cite{xue2017deep} and Bayesian personalized ranking~\cite{rendle2012bpr}) in the post-training setting, as the training data is not available.
As a result, we can only use the regularization term to bound the user embedding with the original one, preventing a drastic change in user embedding.
The idea behind this is that we deem that closer model parameters will lead to closer model performance.
Formally, the regularization loss is defined as $\ell_r = R(\theta, \theta^*)$,
where $\theta^*$ denotes the original user embedding before unlearning.

\subsection{Implementation}\label{sec:imp}
\subsubsection{\textbf{Regularization Loss}} For the regularization loss $\ell_r$, we choose the commonly used regularizer, Frobenius norm~\cite{bottcher2008frobenius}.
Formally,
\begin{equation}
    R(\theta, \theta^*) = \|\theta - \theta^*\|_F^2 = \sum_{i=1}^{N}\sum_{j=1}^K(\theta_{i, j} - \theta^*_{i, j})^2.
\end{equation}

\subsubsection{\textbf{Distinguishability Loss}}
\paragraph{U2U} For the U2U distinguishability loss, we also use Frobenius norm to measure the divergence of user embedding.
Formally, $Div(\theta_i, \theta_j) = \|\theta_i - \theta_j\|^2_F$.
As for attribute divergence, we use the inverse adjacent matrix.
To be specific, 
\begin{equation}\label{equ:div}
    Div(v_i, v_j) = 1 - A(S_i, S_j), \enspace A(S_i, S_j) = \begin{cases}
   1 &\text{if } S_i = S_j.\\
   0 &\text{if } S_j \ne S_j.
\end{cases}
\end{equation}
Due to the property that $Div(v_i, v_j) = 0$ if $S_i = S_j$, we can rewrite U2U loss as:
\begin{equation}
    \ell_{u, U} = \sum_{i=1}^N\sum_{j=1}^N\|\theta_i - \theta_j\|^2_F \cdot Div(v_i, v_j) = 2\text{Tr}(\theta^\top\mathcal{L}_D\theta),
\end{equation}
where $\mathcal{L}_D$ is the Laplacian matrix of divergence $Div(v_i, v_j)$.
This tensorized expression is more efficient for GPU devices to compute.

Although the choice of attribute divergence is not the main focus of this paper, we would like to have a brief discussion about it.
Using the inverse adjacent matrix as attribute divergence is simple and effective, but it loses a certain amount of attribute information and there are more choices with potential.
A straightforward choice is to transform attribute labels into one-hot vectors and apply classification losses, e.g., cross-entropy~\cite{zhang2018generalized} and cosine similarity~\cite{dehak2010cosine}.
Rather than using one-hot vectors, a more advanced approach would be to use the predicted softmax vector from attacking models instead.
This approach can not only preserve more information about attribute labels, but also adopt adversarial training while updating user embedding, which means $Div(v_i, v_j)$ is adaptively updated as $Div(\theta_i, \theta_j)$ is updated.

\paragraph{D2D} For the D2D distinguishability loss, we apply MMD with radial kernels~\cite{tolstikhin2016minimax} to measure the distance of two distributions.
MMD satisfies several properties that are required as a distance measurement, including non-negativity and exchange invariance, i.e., $Dist(P_{\theta_1}, P_{\theta_2}) = Dist(P_{\theta_2}, P_{\theta_1})$.

\subsubsection{\textbf{Summary}} Incorporating different distinguishability losses, we propose two loss functions for the PoT-AU problem.
The U2U-R loss computes as:
\begin{equation}\label{equ:u2u}
    L_{U}(\theta) = \ell_{u, U} + \alpha R(\theta, \theta^*) = 2\text{Tr}(\theta^\top\mathcal{L}_D\theta) + \alpha \|\theta - \theta^*\|_F^2.
\end{equation}
The D2D-R loss computes as:
\begin{equation}\label{equ:d2d}
    L_{D}(\theta) = \ell_{u, D} + \alpha R(\theta, \theta^*) = \text{MMD}(P_{\theta_1}, P_{\theta_2}) + \alpha \|\theta - \theta^*\|_F^2.
\end{equation}
We apply the stochastic gradient descent algorithm~\cite{bottou2012stochastic} to optimize our proposed losses.
\section{Experiments}

To comprehensively evaluate our proposed methods, we conduct experiments on three benchmark datasets and observe the performance in terms of unlearning and recommendation.
We also investigate the efficiency and robustness of our proposed loss functions.
We further analyze the change in user embedding before and after unlearning to better understand the mechanism of our proposed methods.

\subsection{Experimental Settings}

\subsubsection{\textbf{Datasets}}
Experiments are conducted on three publicly accessible datasets that contain both input data, i.e., user-item interactions, and user attributes, i.e., gender.
Following~\cite{ganhor2022unlearning}, the provided gender information of the users are limited to females and males.
\begin{itemize}[leftmargin=*] \setlength{\itemsep}{-\itemsep}
    \item \textbf{MovieLens 100K (ML-100K)}\footnote{https://grouplens.org/datasets/movielens/}: MovieLens is one of the most widely used datasets in the recommendation~\cite{harper2015movielens, he2016ups}.
    They collected users' ratings towards movies as well as other attributes, e.g., gender, age, and occupation.
    ML-100K is the version containing 100 thousand ratings. 
    
    \item \textbf{MovieLens 1M (ML-1M)}: This version has 1 million ratings.
    
    \item \textbf{LFM-2B}\footnote{http://www.cp.jku.at/datasets/LFM-2b}: This dataset collected more than 2 billion listening events, which is used for music retrieval and recommendation tasks~\cite{melchiorre2021investigating}.
    LFM-2B also contains user attributes including gender, age, and country.
\end{itemize} 

We filter out the users and items that have less than 5 interactions. 
Specifically, we use 80\% of ratings for training, 10\% as the validation set for tuning hyper-parameters, and the rest for testing.
Table~\ref{tab:dataset} summarizes the statistics of three datasets.

\begin{table}
\caption{Summary of datasets.}
\label{tab:dataset}
\begin{tabular}{llrrrr}  
\toprule
Dataset & Attribute & User \#   & Item \#   & Rating \# & Sparsity \\
\midrule
\multirow{3}{*}{ML-100K}   & Total & 943     & \multirow{3}{*}{1,682}     & 100,000 & 93.695\%\\
& Female & 273 & & 73,824 & 83.923\%\\
& Male & 670 & & 26,176 & 97.677\%\\
\midrule
\multirow{3}{*}{ML-1M}   & Total & 6040     & \multirow{3}{*}{3,950}     & 1,000,209 & 95.814\%\\
& Female & 1,079 & & 246,896 & 94.207\%\\
& Male & 4,331 & & 753,313 & 95.597\%\\
\midrule
\multirow{3}{*}{LFM-2B}    & Total & 19,972   & \multirow{3}{*}{99,639}   & 2,829,503   & 99.858\%\\
& Female & 4,415 & & 444,076 & 99.899\%\\
& Male & 15,557 & & 2,385,427 & 99.846\%\\
\bottomrule
\end{tabular}
\end{table}

\subsubsection{\textbf{Recommendation Models}}
We test our proposed methods on two different recommendation models.
As mentioned in Section~\ref{sec:rec}, we focus on collaborative filtering models and use the \textit{user embedding}
as the attacking and unlearning target.

\begin{itemize}[leftmargin=*] \setlength{\itemsep}{-\itemsep}
    \item \textbf{NMF}~\cite{he2017neural}: Neural Matrix Factorization (NMF) is one of the representative well-known models based on matrix factorization.
    \item \textbf{LightGCN}~\cite{he2020lightgcn}: Light Graph Convolution Network (LightGCN) is the state-of-the-art collaborative filtering model which improves recommendation performance by simplifying the graph convolution network. 
\end{itemize}


\subsubsection{\textbf{Attackers}}
We randomly expose 10\% user embedding and their corresponding labels as the \textit{shadow dataset} to train attackers, leaving the rest for testing.
We choose two easy-implement and powerful machine learning models as attackers.

\begin{itemize}[leftmargin=*] \setlength{\itemsep}{-\itemsep}
    \item \textbf{MLP}~\cite{gardner1998artificial}: Multilayer Perceptron is a simplified three-layer neural network, which is a widely used classifier.
    \item \textbf{XGB}~\cite{chen2015xgboost}: XGBoost is an acknowledged classifier in the industry. 
    It is so powerful that it has been used in numerous machine learning since it was proposed~\cite{chen2016machine}.
\end{itemize}


\subsubsection{\textbf{Unlearning Methods}} There are lots of studies on machine unlearning, but they are not applicable to the PoT-AU problem.
To the best of our knowledge, we are the first to study the PoT-AU problem.
Although InT-AU setting is different from ours, i.e., PoT-AU, comparing with InT-AU methods would be helpful to provide a comprehensive understanding of the AU problem.
Thus, we compared our proposed U2U-R and D2D-R with the original user embedding and other InT-AU methods.
\begin{table*}
\caption{Results of unlearning performance (performance of attackers). The top results are highlighted in bold.
The lower the attacking performance, the better the unlearning performance.}
\label{tab:unlearn}
\begin{tabular}{lc|rrrr|rrrr}
\toprule
\multicolumn{2}{c}{\multirow{2}{*}{NMF}} & \multicolumn{4}{|c|}{MLP} & \multicolumn{4}{c}{XGB}\\
& & Acc & Precision & Recall & AUC & Acc & Precision & Recall & AUC \\
\midrule
\multirow{5}{*}{ML-100K} & Original & 0.6455 & 0.6735 & 0.5893 & 0.6465 & 0.6364 & 0.6333 & 0.6786 & 0.6356\\
& U2U-R  & \textbf{0.4818} & 0.9997 & 0.0013 & \textbf{0.5002} & 0.9909 & 0.9828 & 0.9999 & 0.9906\\
& D2D-R  & 0.5182 & 0.6333 & 0.3115 & 0.5434 & 0.5091 & 0.5593 & 0.5410 & \textbf{0.5052}\\
& Retrain & 0.5091 & 1.0000 & \textbf{0.0003} & 0.5006 & \textbf{0.3818} & \textbf{0.3965} & \textbf{0.4107} & 0.6187 \\
& Adv-InT & 0.4883 & \textbf{0.5758} & 0.2054 & 0.5135 & 0.5933 & 0.6409 & 0.5662 & 0.5957\\
\midrule
\multirow{5}{*}{ML-1M} & Original & 0.7485 & 0.7403 & 0.7125 & 0.7464 & 0.7310 & 0.7361 & 0.6625 & 0.7269\\
& U2U-R  & \textbf{0.4737} & 1.0000 & \textbf{0.0002} & \textbf{0.5001} & 0.9831 & 0.9763 & 0.9981 & 0.9845\\
& D2D-R & 0.4795 & \textbf{0.4167} & 0.6338 & 0.5019 & 0.5029 & \textbf{0.4186} & 0.5070 & \textbf{0.5035}\\
& Retrain & 0.4883 & 0.4868 & 1.0000 & 0.5028 & \textbf{0.4437} & 0.4598 & \textbf{0.4819} & 0.5261\\
& Adv-InT & 0.4912 & 0.4557 & 0.4161 & 0.5129 & 0.6079 & 0.6551 & 0.5577 & 0.6110\\
\midrule
\multirow{5}{*}{LFM-2B} & Original & 0.7181 & 0.6906 & 0.7501 & 0.7192 & 0.6767 & 0.6419 & 0.7422 & 0.6790 \\
& U2U-R  & \textbf{0.5188} & 0.5641 & \textbf{0.1654} & 0.5188 & 0.9991 & 0.9873 & 0.9924 & 0.9993\\
& D2D-R  & 0.5263 & 0.5128 & 0.6154 & 0.5283 & 0.5075 & 0.4960 & \textbf{0.4769} & 0.5068\\
& Retrain & 0.5226 & \textbf{0.5021} & 0.9531 & 0.5382 & \textbf{0.5038} & \textbf{0.4853} & 0.5156 & \textbf{0.5042} \\
& Adv-InT & 0.5205 & 0.5243 & 0.6207 & \textbf{0.5187} & 0.5825 & 0.6279 & 0.5557 & 0.5812 \\
\midrule
\multicolumn{2}{c}{\multirow{2}{*}{LightGCN}} & \multicolumn{4}{|c|}{MLP} & \multicolumn{4}{c}{XGB}\\
& & Acc & Precision & Recall & AUC & Acc & Precision & Recall & AUC \\
\midrule
\multirow{5}{*}{ML-100K} & Original & 0.6220 & 0.6363 & 0.6512 & 0.6205 & 0.6585 & 0.6596 & 0.7209 & 0.6553\\
& U2U-R & \textbf{0.4391} & 0.9993 & \textbf{0.0012} & \textbf{0.5011} & 0.9756 & 0.95838 & 0.9998 & 0.9722\\
& D2D-R & 0.5244 & 0.4737 & 0.4865 & 0.5210 & 0.5047 & 0.5213 & 0.5132 & \textbf{0.5045}\\
& Retrain & 0.4545 & \textbf{0.4483} & 0.4815 & 0.5449 & \textbf{0.4795} & \textbf{0.4952} & \textbf{0.4962} & 0.5107  \\
& Adv-InT & 0.5375 & 0.4907 & 0.5221 & 0.5326 & 0.5785 & 0.6032 & 0.6007 & 0.5864\\
\midrule
\multirow{5}{*}{ML-1M} & Original & 0.7076 & 0.6871 & 0.6957 & 0.7069 & 0.6754 & 0.6524 & 0.6646 & 0.6748\\
& U2U-R & 0.5175 & 0.5134 & 0.9943 & \textbf{0.5090} & 0.9874 & 0.9761 & 0.9973 & 0.9889\\
& D2D-R & 0.5234 & 0.5281 & 0.5434 & 0.5232 & 0.5146 & 0.5205 & 0.5145 & 0.5146\\
& Retrain & \textbf{0.4737} & \textbf{0.4641} & \textbf{0.4201} & 0.5269 & \textbf{0.5029} & \textbf{0.4969} & \textbf{0.4793} & \textbf{0.5027}\\
& Adv-InT & 0.5117 & 0.4908 & 0.8209 & 0.5272 & 0.5917 & 0.6544 & 0.5725 & 0.6037\\
\midrule
\multirow{5}{*}{LFM-2B} & Original & 0.7218 & 0.6712 & 0.7903 & 0.7261 & 0.6541 & 0.6143 & 0.6935 & 0.6566 \\
& U2U-R & \textbf{0.4812} & 0.4773 & 0.9981 & 0.5071 & 0.9997 & 0.9989 & 0.9874 & 0.9998\\
& D2D-R & 0.5062 & 0.5151 & \textbf{0.4928} & \textbf{0.5064} & \textbf{0.5113} & 0.5286 & \textbf{0.5312} & \textbf{0.5103}\\
& Retrain & 0.5489 & \textbf{0.4247} & 0.5000 & 0.5458 & 0.5263 & \textbf{0.4459} & 0.5323 & 0.5226\\
& Adv-InT & 0.5439 & 0.5139 & 0.5714 & 0.5454 & 0.5879 & 0.6124 & 0.5548 & 0.5734\\
\bottomrule
\end{tabular}
\end{table*}

\begin{itemize}[leftmargin=*] \setlength{\itemsep}{-\itemsep}
    \item \textbf{U2U-R} (PoT-AU): This is the two-component loss function with user-to-user loss as distinguishability loss, i.e., Equation~(\ref{equ:u2u}).
    %
    \item \textbf{D2D-R} (PoT-AU): This is the two-component loss function with distribution-to-distribution loss as distinguishability loss, i.e., Equation~(\ref{equ:d2d}).
    \item \textbf{Original}: This is the original user embedding before unlearning.
    \item \textbf{Retrain}~\cite{zafar2019fairness} (InT-AU): This method adds a regularizer to the original loss to achieve fairness. Inspired by it, we add our proposed D2D loss to the original recommendation loss and retrain the model from scratch.
    Note that we only use D2D loss because incorporating U2U loss in this setting would be computationally prohibitive.
    \item \textbf{Adv-InT}~\cite{ganhor2022unlearning} (InT-AU): This method uses adversarial training to achieve InT-AU for variational auto-encoder.
    We also apply the idea of adversarial training to our tested recommendation models, i.e., NMF and LightGCN, and name it Adv-InT.
\end{itemize}


We run all models 10 times and report the average results.
Due to the space limit, we report the hyper-parameter settings and hardware information in Appendix~\ref{sec:app_exp}.

\begin{table}
\caption{Results of recommendation performance. The top results are highlighted in bold.}
\label{tab:rec}
\resizebox{\linewidth}{!}{
\begin{tabular}{lc|rrrr}
\toprule
\multicolumn{2}{c}{NMF} & NDCG@5 & HR@5 & NDCG@10 & HR@10 \\
\midrule
\multirow{5}{*}{ML-100K} & Original  & 0.4308 & 0.6098 & \textbf{0.4179} & \textbf{0.4491} \\
& U2U-R & 0.4174 & 0.6015 & 0.4073 & 0.4425 \\
& D2D-R & \textbf{0.4320} & \textbf{0.6102} & 0.4147 & 0.4489 \\
& Retrain & 0.4159 & 0.6004 & 0.4109 & 0.4439\\
& Adv-InT & 0.4213 & 0.5997 & 0.4097 & 0.4453\\
\midrule
\multirow{5}{*}{ML-1M} & Original  & \textbf{0.5146} & \textbf{0.7517} & \textbf{0.5054} & \textbf{0.5761} \\
& U2U-R & 0.4316 & 0.6957 & 0.4430 & 0.5443 \\
& D2D-R & 0.5110 & 0.7497 & 0.5033 & 0.5750 \\
& Retrain & 0.5095 & 0.7421 & 0.4962 & 0.5673 \\
& Adv-InT & 0.4913 & 0.7395 & 0.4875 & 0.5514\\
\midrule
\multirow{5}{*}{LFM-2B} & Original  & 0.2082 & \textbf{0.8510} & \textbf{0.3773} & \textbf{0.8130} \\
& U2U-R & 0.1342 & 0.6572 & 0.2268 & 0.6236 \\
& D2D-R & \textbf{0.2111} & 0.8479 & 0.3746 & 0.8103 \\
& Retrain & 0.2033 & 0.8414 & 0.3695 & 0.8071 \\
& Adv-InT & 0.2012 & 0.8335 & 0.2706 & 0.8075 \\
\midrule
\multicolumn{2}{c}{LightGCN} & NDCG@5 & HR@5 & NDCG@10 & HR@10\\
\midrule
\multirow{5}{*}{ML-100K} & Original  & \textbf{0.4394} & \textbf{0.6161} & \textbf{0.4195} & \textbf{0.4496} \\
& U2U-R & 0.4274 & 0.6106 & 0.4079 & 0.4418 \\
& D2D-R & 0.4384 & 0.6131 & 0.4180 & 0.4492 \\
& Retrain & 0.4211 & 0.6051 & 0.4092 & 0.4450 \\
& Adv-InT & 0.4285 & 0.6063 & 0.4134 & 0.4443\\
\midrule
\multirow{5}{*}{ML-1M} & Original  & \textbf{0.4554} & \textbf{0.7170} & \textbf{0.4674} & \textbf{0.5590} \\
& U2U-R & 0.4296 & 0.6983 & 0.4445 & 0.5452 \\
& D2D-R & 0.4534 & 0.7168 & 0.4657 & 0.5586 \\
& Retrain & 0.4471 & 0.7176 & 0.4666 & 0.5567 \\
& Adv-InT & 0.4485 & 0.7106 & 0.4625 & 0.5538\\
\midrule
\multirow{5}{*}{LFM-2B} & Original  & \textbf{0.2234} & \textbf{0.8898} & \textbf{0.4025} & \textbf{0.8471}\\
& U2U-R & 0.2148 & 0.8807 & 0.3898 & 0.8332 \\
& D2D-R & 0.2228 & 0.8894 & 0.4018 & 0.8461 \\
& Retrain & 0.2230 & 0.8885 & 0.3996 & 0.8449 \\
& Adv-InT & 0.2213 & 0.8827 & 0.3943 & 0.8425\\
\bottomrule
\end{tabular}
}
\end{table}


\begin{table}
\caption{Running time of unlearning methods.}
\label{tab:time}
\begin{tabular}{lc|rrrr}
\toprule
\multicolumn{2}{c}{Time (s)} & U2U-R & D2D-R & Retrain & Adv-InT\\
\midrule
\multirow{2}{*}{ML-100K} & NMF  & 5.78 & 2.69 & 117.39 & 223.21\\
& LightGCN  & 11.86 & 5.99 & 284.65 & 567.24\\
\midrule
\multirow{2}{*}{ML-1M} & NMF  & 66.65 & 28.80 & 629.09 & 928.55\\
& LightGCN  & 159.22 & 60.99 & 1402.22 & 2103.57\\
\midrule
\multirow{2}{*}{LFM-2B} & NMF  & 110.05 & 45.00 & 1116.19 & 1457.31\\
& LightGCN  & 364.73 & 147.54 & 2993.87 & 3503.16\\
\bottomrule
\end{tabular}
\end{table}



\subsection{Results and Discussions}
%

\subsubsection{\textbf{Unlearning Performance}}
The performance of unlearning is evaluated by the performance of attackers.
We train attackers on the shadow training set, and report their performance on the testing set. 
To comprehensively evaluate attacking performance, we report four metrics, including Accuracy (Acc), precision, recall, and Area Under the ROC Curve (AUC)~\cite{fawcett2006introduction}, in Table~\ref{tab:unlearn}.
To visually analyze the results, we also use t-SNE~\cite{van2008visualizing} to reduce dimension (Figure~\ref{fig:tsne} in Appendix~\ref{sec:app_tsne}).
%
%
We have the following observations from the above results. 
Firstly, attackers achieve an average accuracy of 0.68 on the original embedding, indicating that information on the user's attribute in user embedding is released to the attackers.
Secondly, all methods can degrade the attacking performance of MLP.
U2U-R, D2D-R, Retrain, and Adv-InT decrease the AUC by 27.11\%, 25.00\%, 24.16\%, and 24.37\%, respectively, on average.
%
%
Thirdly, U2U-R cannot fool XGB and increase the attacking performance significantly instead (average AUC is up to 0.99). 
We will further analyze U2U-R's different performance w.r.t MLP and XGB in Section~\ref{sec:emb}.
D2D-R can decrease the AUC of XGB by 24.41\% on average. 
In comparison, Retrain and Adv-InT can only decrease the AUC by 20.93\% and 11.84\%, respectively.

\nosection{Summary} Compared with U2U-R, D2D-R is more effective in unlearning.
D2D-R protects the user's attributes by making them indistinguishable to the attacker, outperforming the compared methods.

\subsubsection{\textbf{Recommendation Performance}}
We use two common metrics, i.g., Normalized Discounted Cumulative Gain (NDCG) and Hit Ratio (HR), to evaluate recommendation performance~\cite{he2015trirank,xue2017deep}.
%
We truncate the ranked list at 5 and 10 for both metrics.
As shown in Table~\ref{tab:rec}, unlearning methods also affect recommendation performance.
Compared with the original performance, U2U-R has an average degradation of 5.30\% on NDCG and 3.77\% on HR.
We analyze the reasons for degradation in Section~\ref{sec:emb}.
Interestingly, D2D-R increases the performance at an average of 6.28\% on NDCG and 1.89\% on HR.
D2D loss, which is devised to make attributes (gender) indistinguishable, could accidentally diminish the negative gender discrimination to enhance recommendation performance.

\nosection{Summary} U2U-R degrades recommendation performance to some extent, while D2D-R outperforms all compared methods and even slightly outperforms the original user embedding.

\subsubsection{\textbf{Efficiency}}
We use running time to evaluate the efficiency of unlearning methods.
From Table~\ref{tab:time}, we observe that i) our proposed PoT-AU methods (U2U-R and D2D-R) significantly outperform InT-AU methods (Retrain and Adv-InT).
This is because PoT-AU methods can be viewed as a fine-tuning process on an existing model, providing them with inherent efficiency compared to InT-AU methods;
ii) By incorporating our proposed distinguishability loss to the original recommendation loss and retraining from scratch, Retrain outperforms Adv-InT.
By serving as a baseline method, Retrain provides a new path for InT-AU methods to explore.

\subsubsection{\textbf{Effect of Trade-off Parameter}}
To investigate the robustness of our proposed two-component loss function, we study the effect of $\alpha$, i.e., trade-off parameter. 
As shown in Figure~\ref{fig:alpha} (Appendix~\ref{sec:app_alpha}), we use AUC and NDCG@10 to represent unlearning and recommendation performance respectively.
We observe that our proposed methods, especially U2U-R, are robust with different $\alpha$.
Reducing the value of $\alpha$ results in insignificant performance change for D2D-R, i.e., enhances unlearning performance and decreases recommendation performance.

\subsubsection{\textbf{Analysis of Embedding}}\label{sec:emb}
To understand the mechanism of our proposed methods and compare the difference between two types of distinguishability losses, we analyze the distributions of user embedding.
Specifically, we report the histograms for each dimension of user embedding where the users are grouped by gender.
Figure~\ref{fig:mat} illustrates the user embedding of LightGCN trained on different datasets.
From it, we have the following observations.

\begin{itemize}[leftmargin=*] \setlength{\itemsep}{-\itemsep}
    \item Compared with original user embedding, U2U-R and D2D-R both change the distributions of user embedding. 
    According to numeric results, i.e., Tables \ref{tab:unlearn} and \ref{tab:rec}, these changes affect unlearning and recommendation performance.
    \item U2U loss does make users with different gender behave more similarly.
    %
    %
    However, U2U-R enhances XGB's attacking performance, instead of degrading it.
    We observe that the current implementation of U2U-R bounds the loss so tightly that it destroys the original distributions, making them collapse to the mean of the original embedding.
    This results in two \textit{needle-shaped} distributions, which makes MLP difficult to distinguish them.
    As for XGB, it has a powerful fitting ability and is sometimes over-fitting.
    But in this case, XGB can over-fit the mean in the training set, and accurately distinguish needle-shaped distributions in the testing set.
    This observation also provides a reasonable explanation for U2U-R's performance drop in the recommendation task (Table~\ref{tab:rec}).
    \item D2D loss is effective in enlarging the overlapping area of two gender distributions, which means it narrows the distance between two distributions. 
    At the same time, it does not deform the original distributions.
    This brings superior performance in both unlearning and recommendation.
\end{itemize}

\nosection{Summary} With the analysis of embedding distribution, we find that the performance degradation of U2U-R in the recommendation task is caused by a sharp change in embedding distribution.
U2U distinguishability loss is bounded so tightly that it makes all users within the distribution move toward the mean.
\section{Related Work}

\subsection{Machine Unlearning}
Machine unlearning aims to remove the influence of particular training data
on a learned model~\cite{nguyen2022survey}.
%
%
%
Existing unlearning methods can be classified into the following two approaches.

\nosection{Exact Unlearning} This approach aims to ensure that the target is unlearned as completely as retraining from scratch.
Cao and Yang~\cite{cao2015towards} transformed training data points into a reduced number of summations to enhance unlearning efficiency. 
%
%
Recently, Bourtoule~et~al.~\cite{bourtoule2021machine} proposed a partition-aggregation unlearning framework, i.e. SISA, which partitions the dataset into disjoint subsets, trains one model on each subset, and aggregates all models.
This design reduces the retraining overhead to subsets.

\nosection{Approximate Unlearning} This approach aims to estimate the influence of unlearning target, and directly remove the influence through parameter manipulation, i.e., updating parameters with the purpose of unlearning~\cite{golatkar2020eternal,guo2020certified,sekhari2021remember,warnecke2021machine}.
%
%
In this approach, the influence of unlearning target is evaluated by influence function~\cite{koh2017understanding, koh2019accuracy}, which is found to be fragile in deep learning~\cite{basu2021influence}.
Recent studies also point out that the influence of individual training data on deep models is intractable to compute analytically~\cite{graves2021amnesiac}.

\subsection{Recommendation Unlearning}
RecEraser was proposed to achieve unlearning in recommender systems~\cite{chen2022recommendation}.
Following SISA's partition-aggregation framework, RecEraser groups similar data together, instead of random partitioning.
%
%
In addition, RecEraser uses an attention-based aggregator to further enhance performance.
Similarly, LASER also groups similar data together~\cite{li2022making}.
However, instead of parallel training, LASER employs sequential training.
While this modification significantly enhances model utility, it comes at the cost of reduced efficiency.
%
Lately, approximate unlearning is also investigated in recommendation to enhance efficiency~\cite{li2023selective}.

\subsection{Attribute Unlearning}
Existing studies predominately focus on unlearning the input data samples, ignoring the latent attributes that are irrelevant to the training process.
AU is first studied by~\cite{guo2022efficient} to unlearn particular attributes of a facial image, e.g., smiling, big nose, and mustache, by adding network blocks.
As recommender systems potentially capture users' sensitive information, e.g., gender, race, and age, it is non-trivial to study AU in the recommendation scenario.

However, splitting the model into a feature extractor and a classifier, and adding a network block between them may not be universally applicable in the context of recommendation~\cite{guo2022efficient}.
Adversarial training was used to achieve AU in recommender systems~\cite{ganhor2022unlearning}.
However, it is under the setting of In-Training AU (InT-AU), which manipulates the model parameters during training.
Different from InT-AU setting, our PoT-AU setting is more strict and practical because 
i) we can only manipulate the model parameters when training is completed, 
ii) as the training data or other training information, e.g., gradients, are usually protected or discarded after training, we cannot get access to them to enhance performance, 
and iii) it is more flexible for recommendation platforms to manipulate the model based on unlearning requests without interfering with the original training process.
\section{Conclusions}\label{sec:con}

In this paper, we study the Post-Training Attribute Unlearning (PoT-AU) problem in recommender systems, which aims to protect users' attribute information instead of input data.
To the best of our knowledge, we are the first to study this problem, which is more strict and practical than In-Training Attribute Unlearning (InT-AU) problem.
There are two goals in the PoT-AU problem, i.e., making attributes indistinguishable, and maintaining comparable recommendation performance.
To achieve the above two goals, we propose a two-component loss function, which consists of distinguishability loss and regularization loss, to optimize user embedding.
We design two types of distinguishability losses from different perspectives, i.e., User-to-User (U2U) and Distribution-to-Distribution (D2D).
We conduct extensive experiments on three real-world datasets to evaluate the effectiveness of our proposed methods, i.e., U2U-R and D2D-R.
Results indicate that i) both methods achieved the unlearning goal, but D2D-R significantly outperformed U2U-R, and ii) D2D-R had a negative impact on recommendation performance, but D2D-R can enhance it.
Generally speaking, D2D-R achieves both goals in the PoT-AU problem.

In this work, we focus on the attribute with binary labels.
%
In the future, we will study the multiple-labels case and further exploit different implementations of the current U2U design.
One of the advantages of U2U is that it can be directly generalized to the multiple-label case without modification.

\begin{acks}
This work was supported in part by the ``Ten Thousand
Talents Program'' of Zhejiang Province for Leading Experts
(No. 2021R52001), and the National Natural Science Foundation of China (No. 72192823).
\end{acks}

\bibliographystyle{ACM-Reference-Format}
\bibliography{main}

\appendix
\begin{figure}[t]
    \centering
    \includegraphics[width=0.8\linewidth]{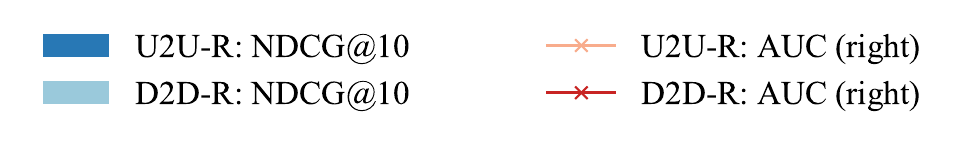}\\
    \vspace{-0.2cm}
    \subfigure[ML-100K: NMF]{
        \includegraphics[height=2.cm]{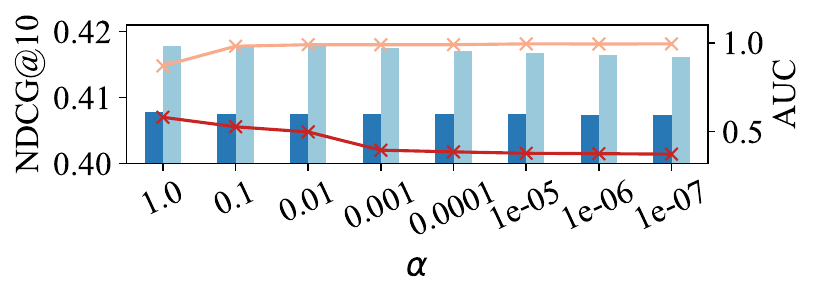}
    }
    \subfigure[ML-1M: NMF]{
        \includegraphics[height=2.cm]{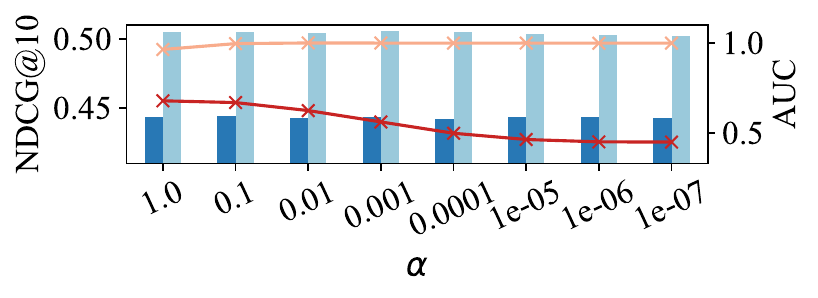}
    }
    \subfigure[LFM-2B: NMF]{
        \includegraphics[height=2.cm]{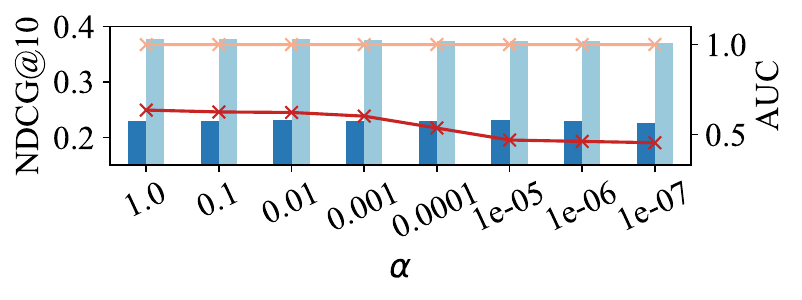}
    }
    \subfigure[ML-100K: LightGCN]{
        \includegraphics[height=2.cm]{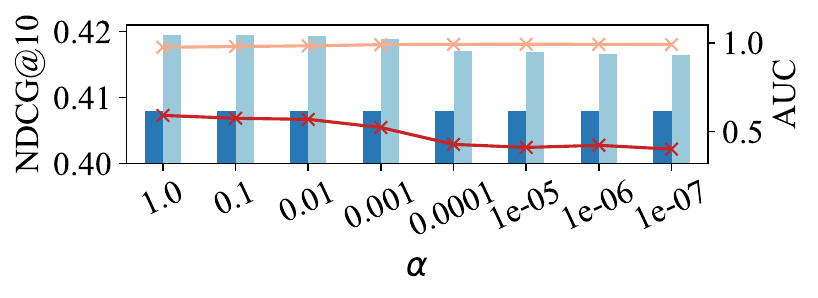}
    }
    \subfigure[ML-1M: LightGCN]{
        \includegraphics[height=2.cm]{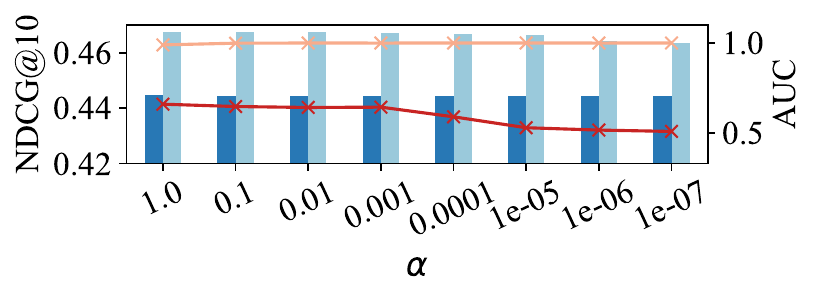}
    }
    \subfigure[LFM-2B: LightGCN]{
        \includegraphics[height=2.cm]{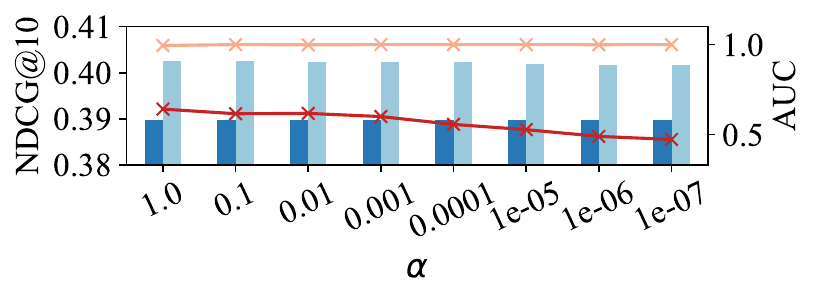}
    }
    \caption{Effect of $\alpha$ w.r.t. unlearning (AUC) and recommendation (NDCG@10) performance.}
    \Description{A flow chart.}
    \label{fig:alpha}
\end{figure}

\begin{figure}[t]
    \centering
    \subfigure[U2U-R: GT]{\label{fig:tsne_gt}
        \includegraphics[width=2.6cm]{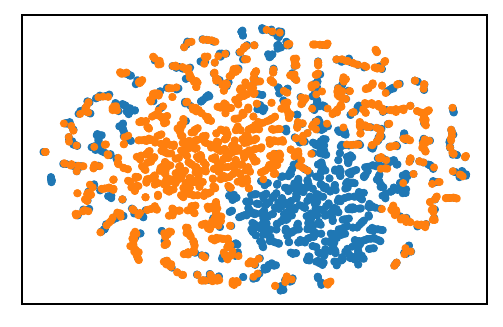}
    }
    \subfigure[U2U-R: MLP]{
        \includegraphics[width=2.6cm]{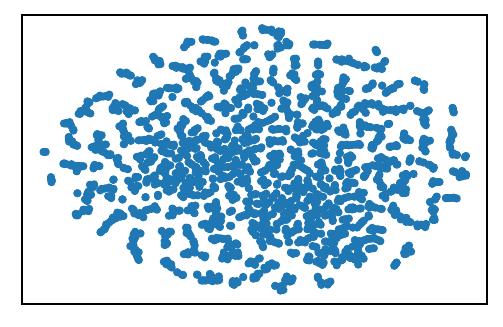}
    }
    \subfigure[U2U-R: XGB]{\label{fig:tsne_xgb}
        \includegraphics[width=2.6cm]{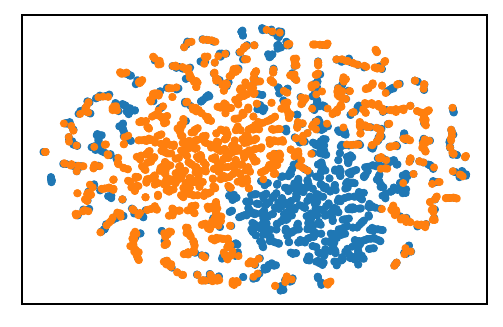}
    }
    \subfigure[D2D-R: GT]{
        \includegraphics[width=2.6cm]{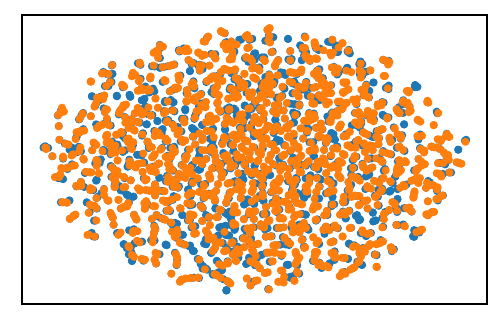}
    }
    \subfigure[D2D-R: MLP]{
        \includegraphics[width=2.6cm]{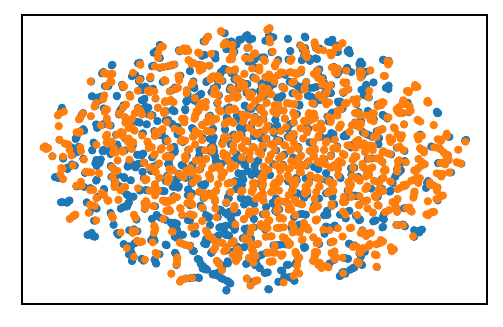}
    }
    \subfigure[D2D-R: XGB]{
        \includegraphics[width=2.6cm]{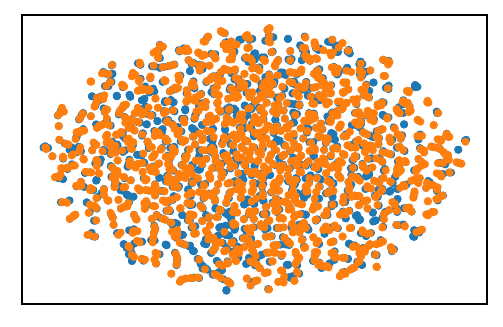}
    }
    \caption{Gender prediction of different user embedding where GT denotes ground truth, the orange and blue dots represent the predictions of female and male respectively.
    The user embedding is trained in LightGCN on ML-1M.}
    \Description{A flow chart.}
    \label{fig:tsne}
\end{figure}
\section{Experimental Settings}\label{sec:app_exp}

\nosection{Hardware Information} All models and algorithms are implemented with Python 3.8 and PyTorch 1.9.
We run all experiments on an Ubuntu 20.04 LTS System server with 48-core CPU, 256GB RAM and NVIDIA GeForce RTX 3090 GPU.

\nosection{Recommendation Models} To obtain the optimal performance, we use grid search to tune the hyper-parameters.
For model-specific hyper-parameters, we follow the suggestions from their original papers.
All model parameters are initialized with a Gaussian distribution $\mathcal{N}(0, 0.01^2)$.
Specifically, we set the learning rate to 0.001 and the embedding size $K$ to 16.
The number of epochs is set to 50 for NMF and 400 for LightGCN.

\nosection{Attackers} For MLP, we set the L2 regularization weight to 1.0 and the maximal iteration to 1000, leaving the other hyper-parameters at their defaults in scikit-learn 1.1.3.
For XGB, we use the xgboost package, setting the hyper-parameters as their default values. 

\nosection{Our Methods} We set $\alpha$ and the learning rate to 1e-4 and 0.001 respectively for both methods.
The number of epochs is set to 5,000 and 1,000 for U2U-R and D2D-R respectively.

\section{Experimental Results}

\subsection{Effect of Trade-off Parameter} \label{sec:app_alpha}
The effect of $\alpha$ is reported in Figure~\ref{fig:alpha}.

\subsection{Visualization of Unlearning Performance}\label{sec:app_tsne} To visually analyze the results, we also reduce the embedding dimension to 2 using t-SNE~\cite{van2008visualizing} and plot the distribution of gender prediction in Figure~\ref{fig:tsne}.
Recall that the aim of unlearning is to degrade the performance of attackers.
From Figure~\ref{fig:tsne}, we observe that U2U-R displays a significant degree of diversity towards different attackers, i.e., MLP and XGB.
\begin{itemize}[leftmargin=*] \setlength{\itemsep}{-\itemsep}
    \item For MLP attack, U2U-R fools the attacker by flipping all labels into the same class, which significantly changes the distribution of user embedding.
    This observation can also be illustrated in Figure~\ref{fig:mat}.
    \item For XGB attack, U2U-R cannot degrade the attacking performance.
    The visualized outcome of U2U-R, i.e., Figure~\ref{fig:tsne_xgb}, depicts minimal disparity from the actual ground truth, i.e., Figure~\ref{fig:tsne_gt}.
\end{itemize}


\subsection{Analysis of Embedding}\label{sec:app_emb}
\begin{figure*}[t]
    \centering
    \subfigure[ML-100K: Original user embedding]{
        \includegraphics[width=0.32\textwidth]{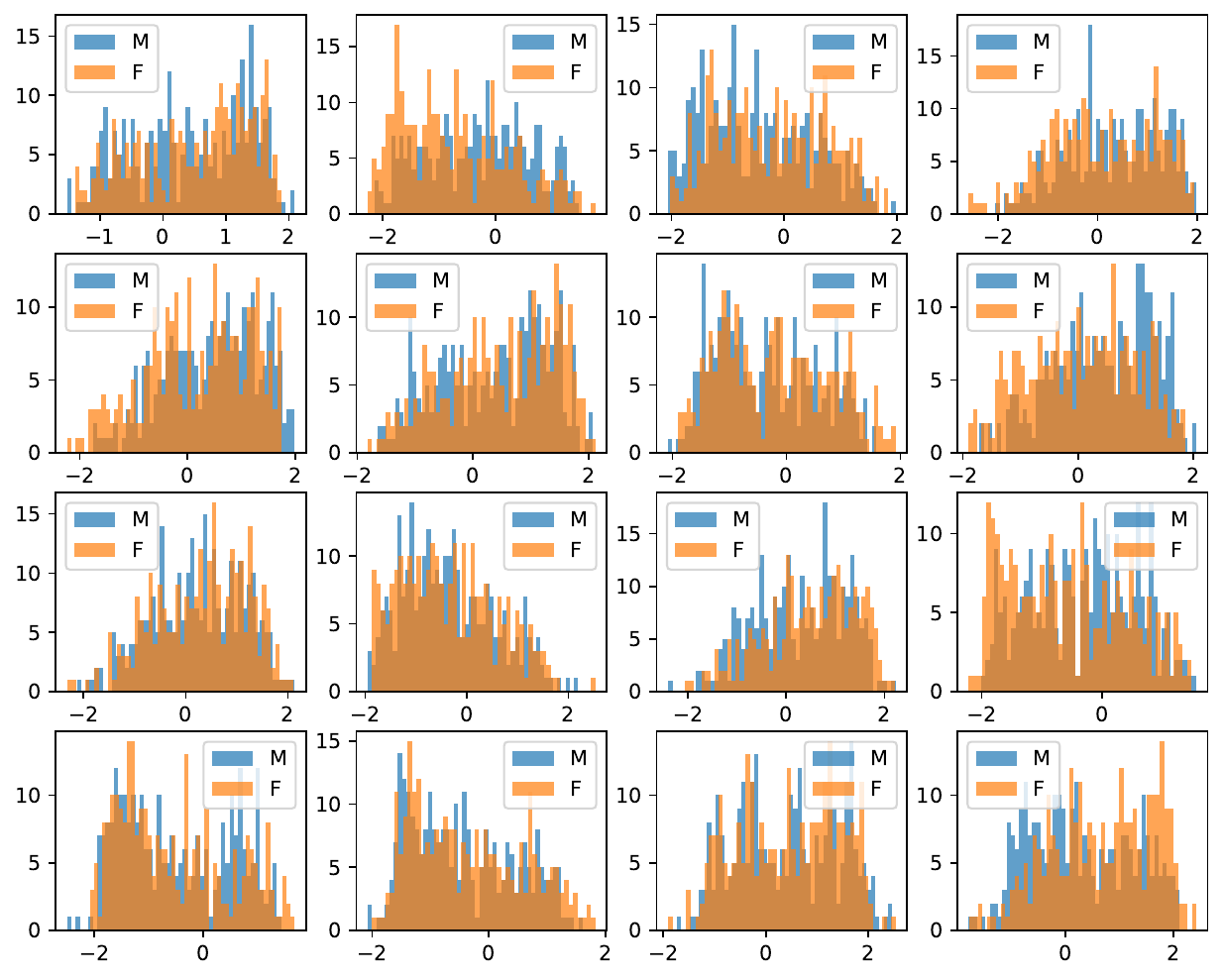}
    }
    \subfigure[ML-100K: U2U-R user embedding]{
        \includegraphics[width=0.325\textwidth]{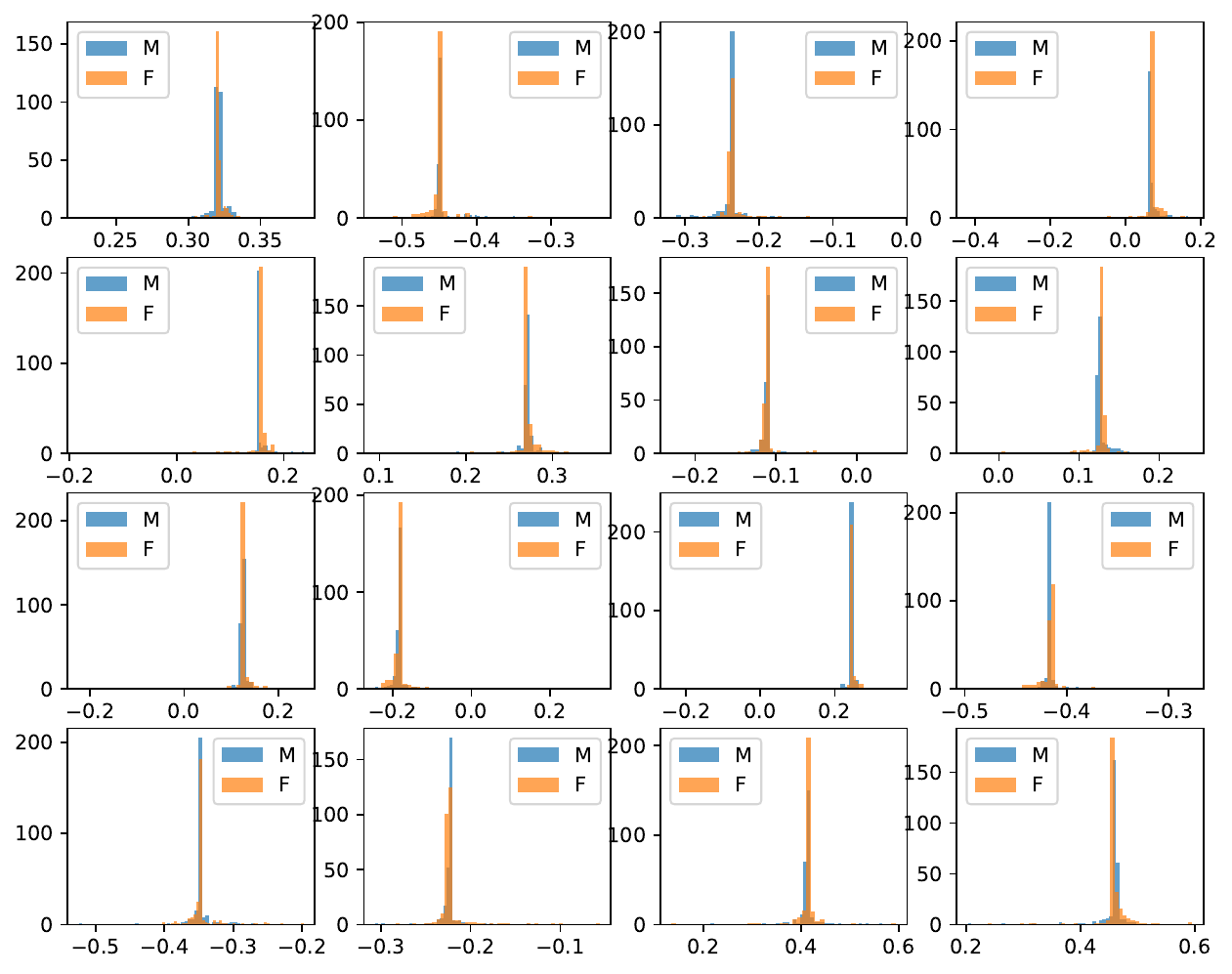}
    }
    \subfigure[ML-100K: D2D-R user embedding]{
        \includegraphics[width=0.32\textwidth]{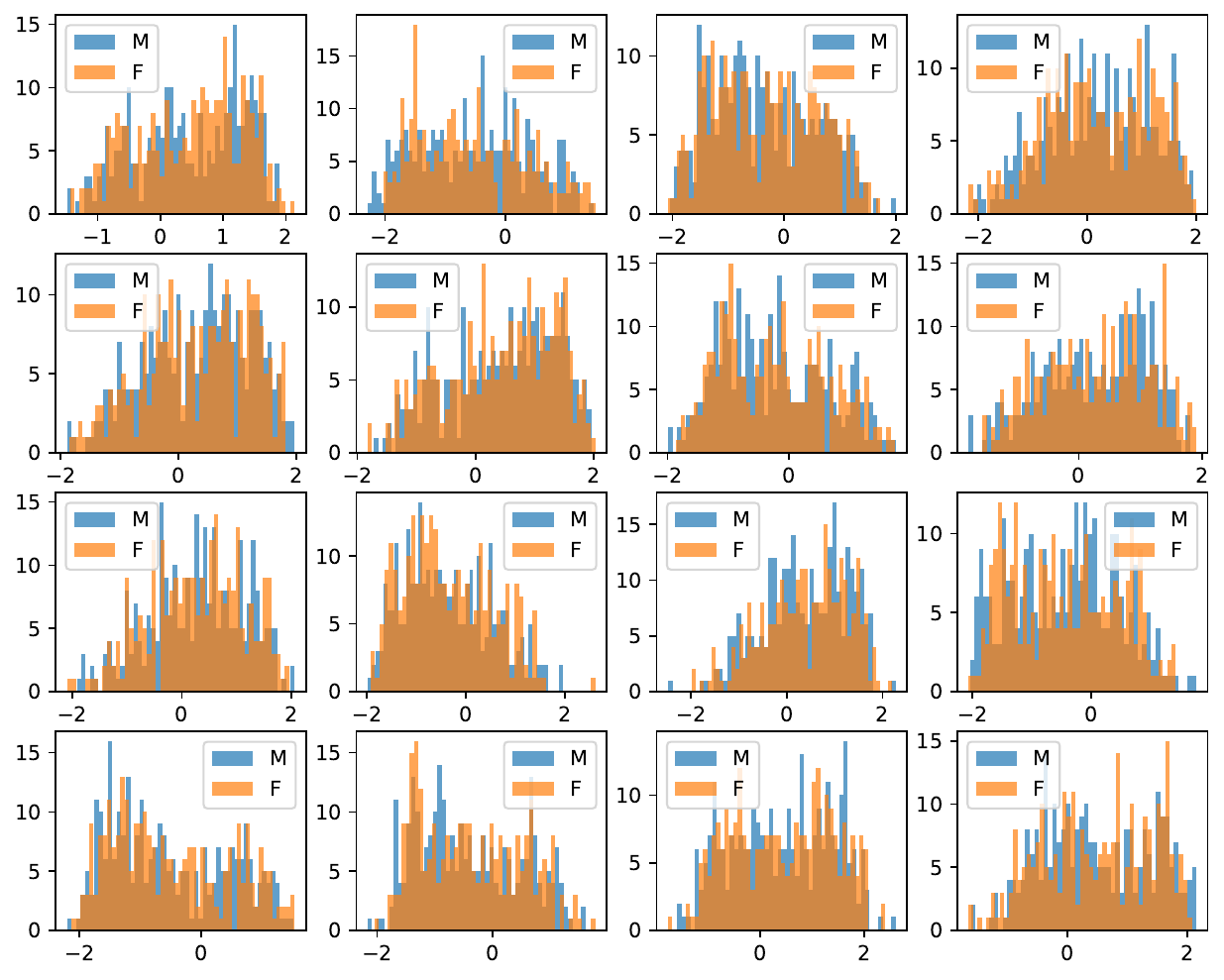}
    }
    \subfigure[ML-1M: Original user embedding]{
        \includegraphics[width=0.32\textwidth]{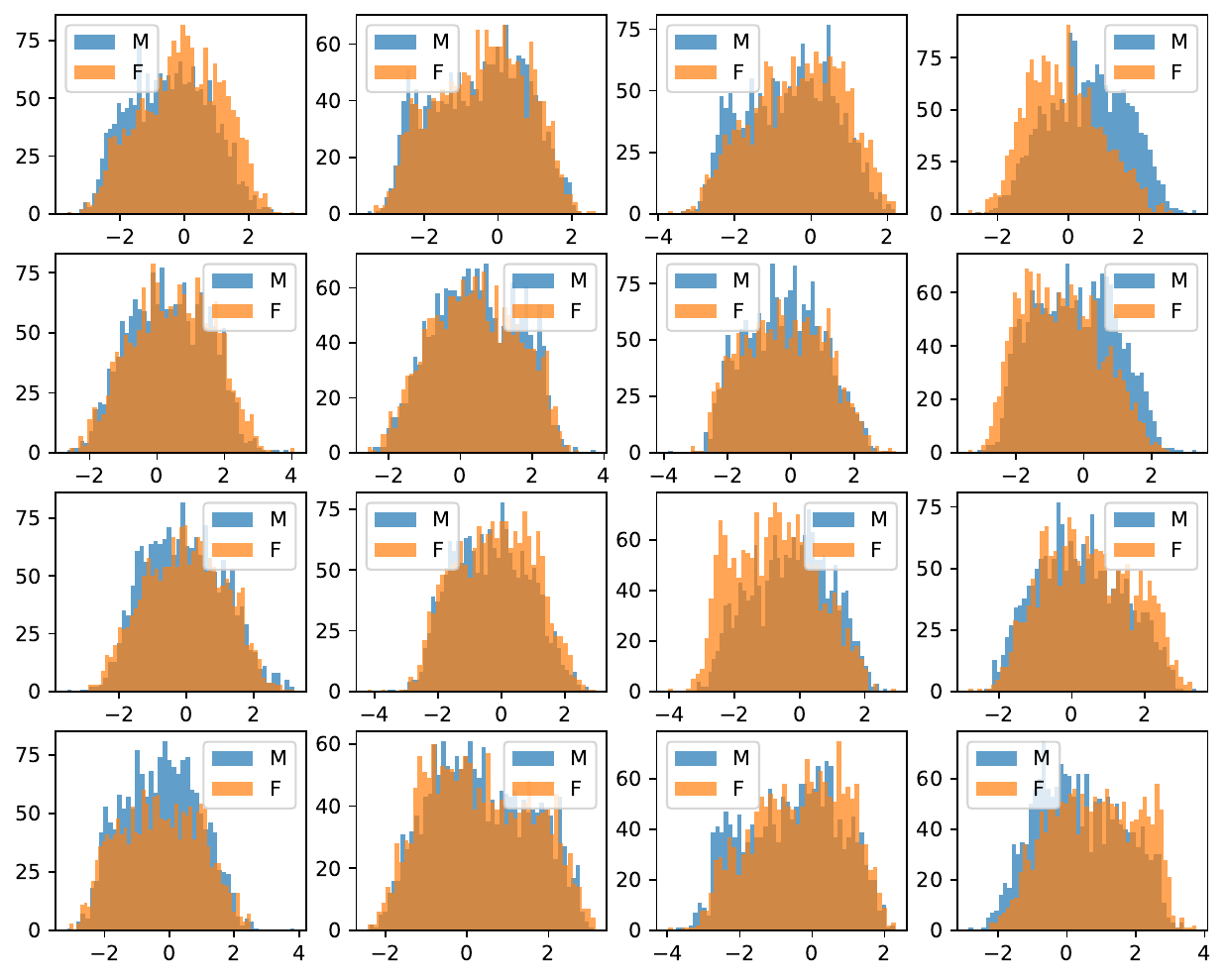}
    }
    \subfigure[ML-1M: U2U-R user embedding]{
        \includegraphics[width=0.325\textwidth]{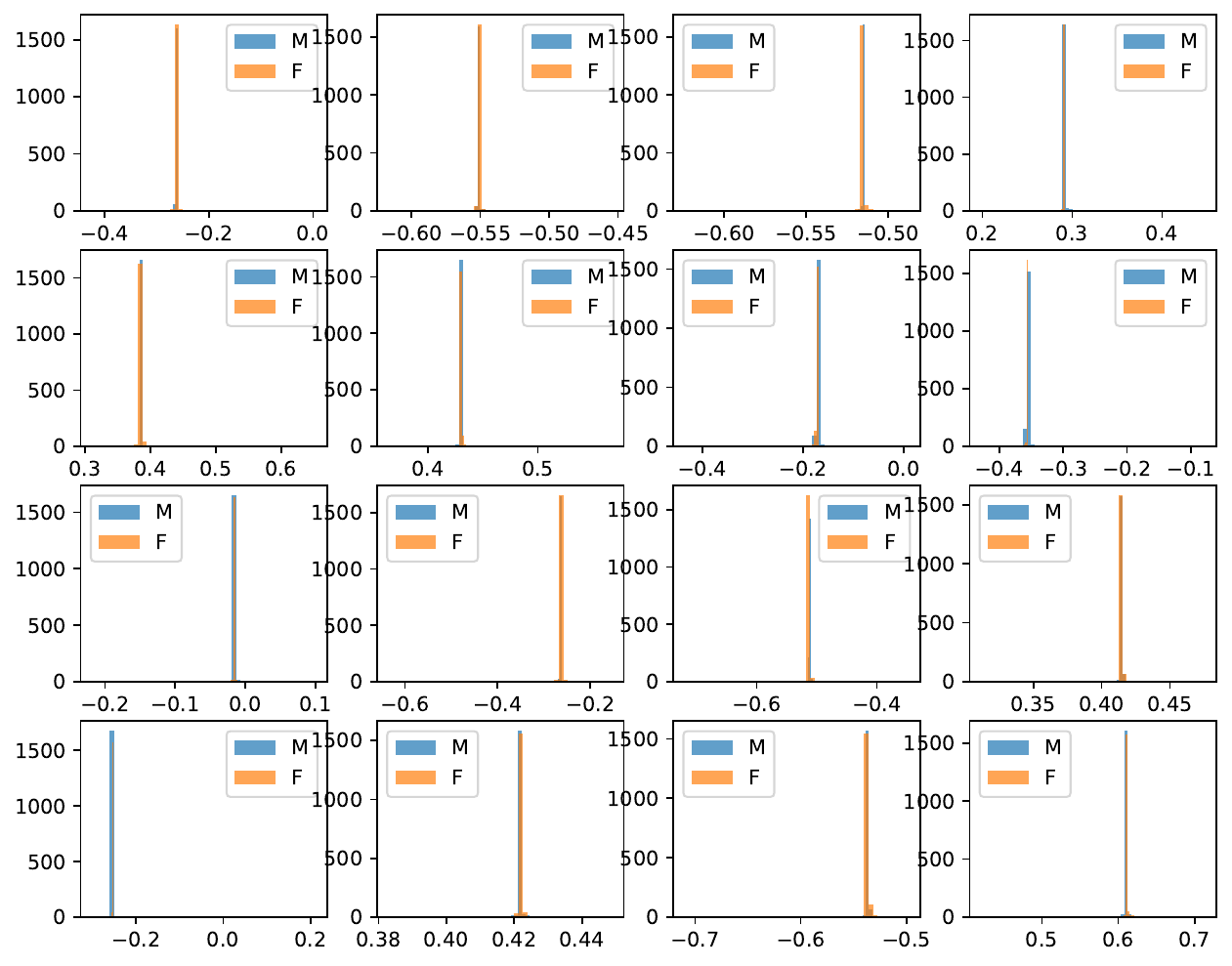}
    }
    \subfigure[ML-1M: D2D-R user embedding]{
        \includegraphics[width=0.32\textwidth]{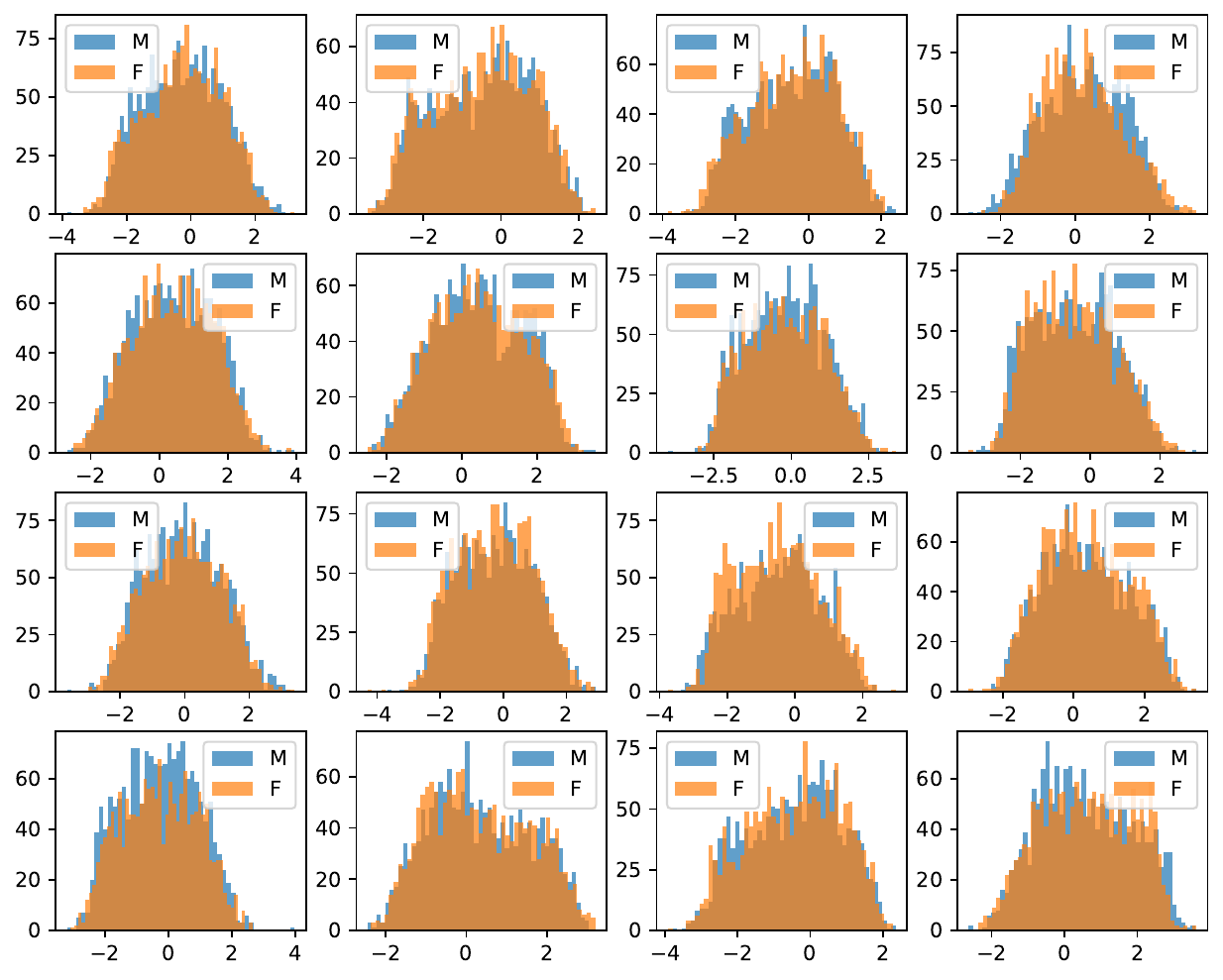}
    }
    \subfigure[LFM-2B: Original user embedding]{
        \includegraphics[width=0.32\textwidth]{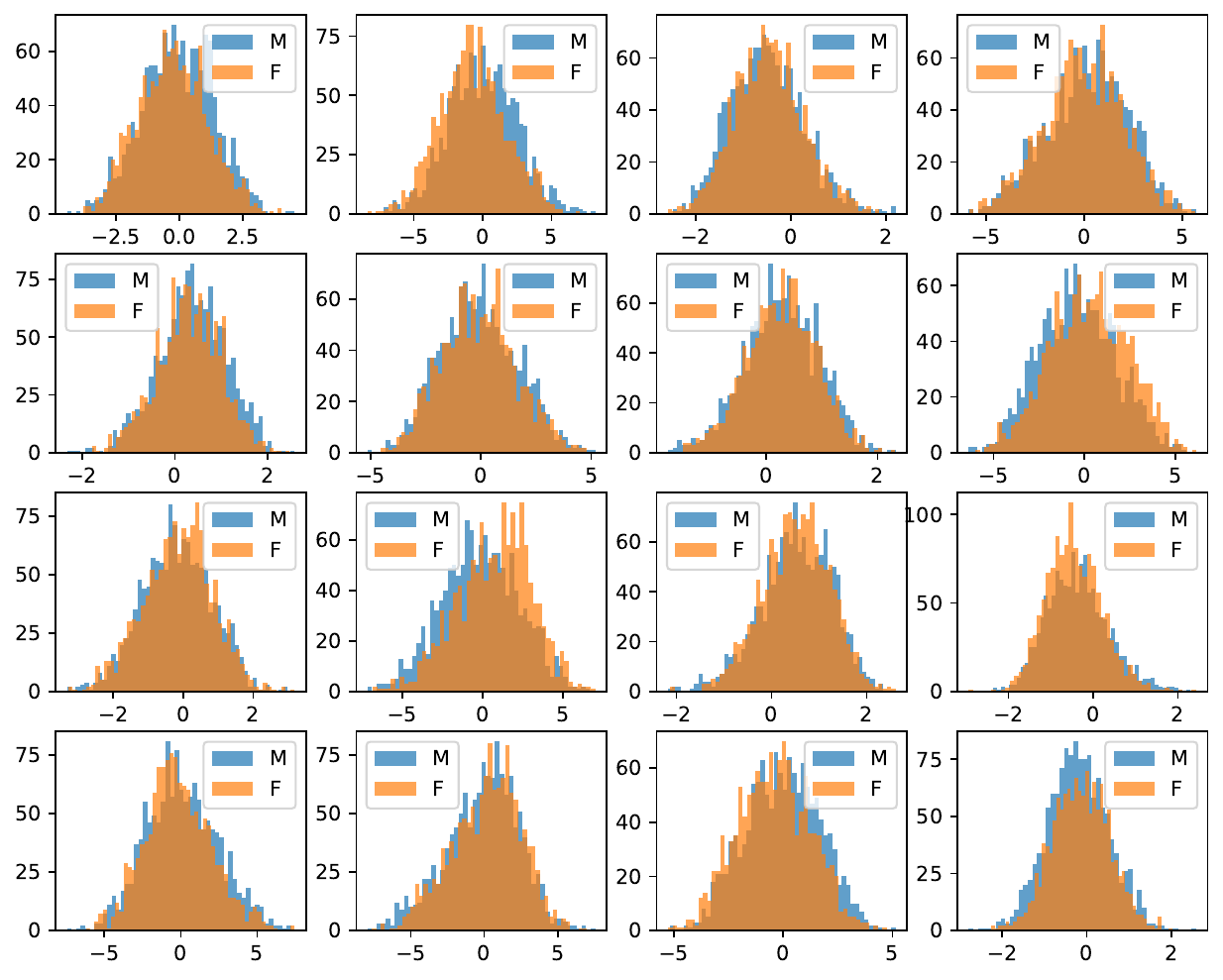}
    }
    \subfigure[LFM-2B: U2U-R user embedding]{
        \includegraphics[width=0.325\textwidth]{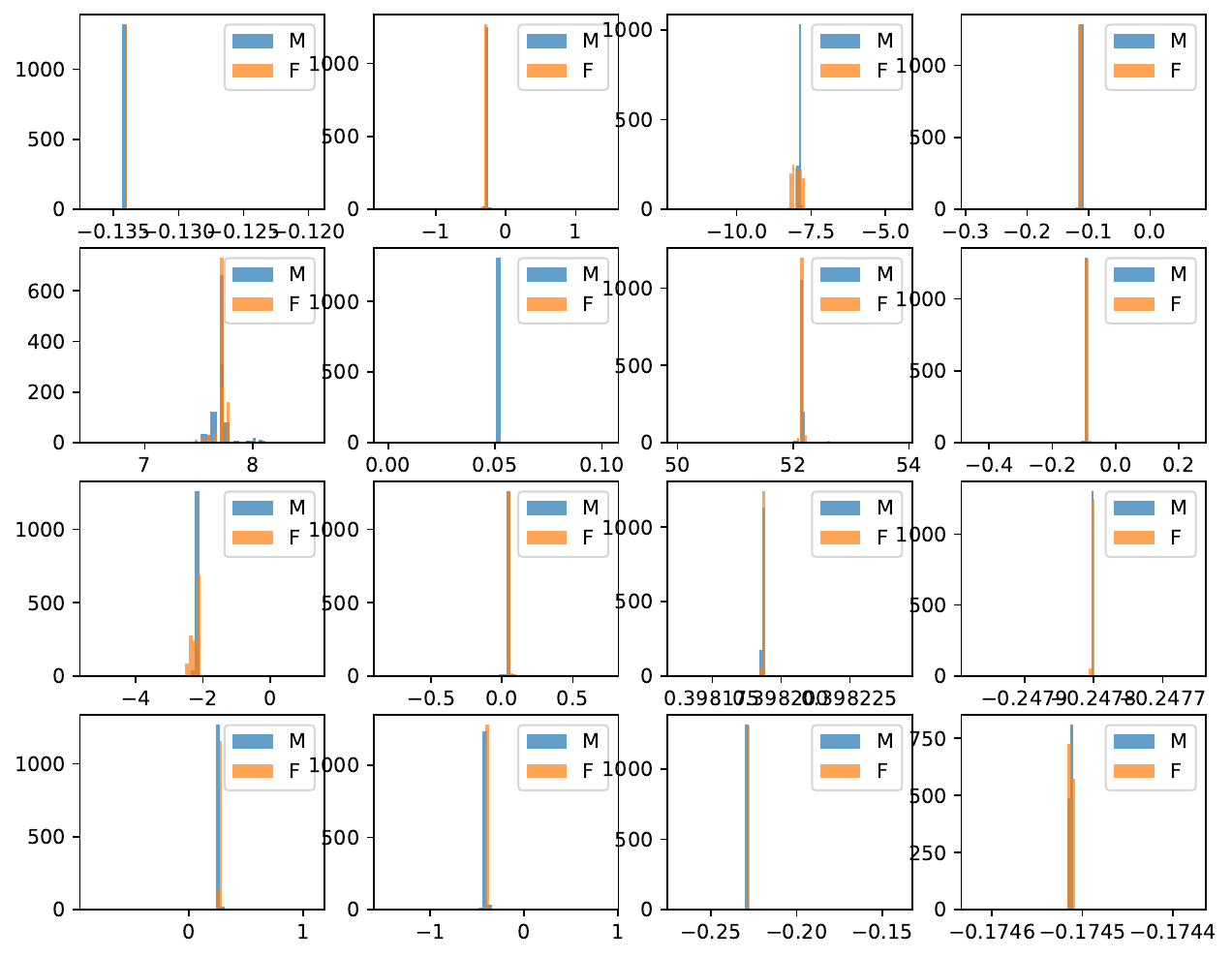}
    }
    \subfigure[LFM-2B: D2D-R user embedding]{
        \includegraphics[width=0.32\textwidth]{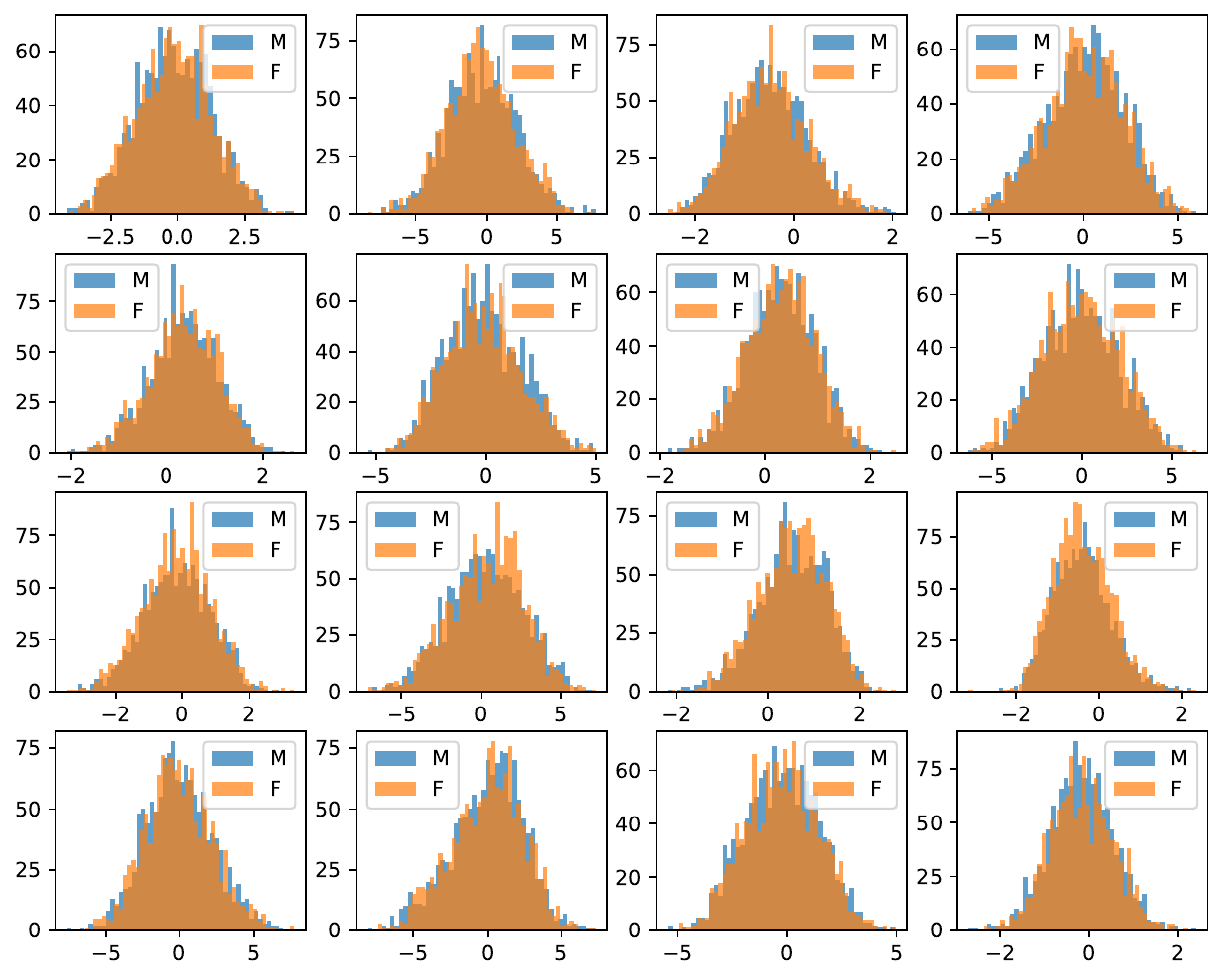}
    }
    \caption{The distributions of different user embedding on different datasets (LightGCN). 
    Each mini-plot represents one dimension of user embedding where F and M denote female and male respectively.
    To compare the two distributions, i.e., female and male, more accurately, we down-sample the males so that the number of males is equal to the number of females.
    }
    \Description{A flow chart.}
    \label{fig:mat}
\end{figure*}

The histograms of user embedding are reported in Figure~\ref{fig:mat}.

\end{document}